\documentclass[10pt, conference, compsocconf]{IEEEtran}
\pdfoutput=1

\usepackage{cite}
\usepackage{ifpdf}
\usepackage{tabularx}
\usepackage[table]{xcolor}
\newcolumntype{L}[1]{>{\hsize=#1\hsize\raggedright\arraybackslash}X}%
\newcolumntype{R}[1]{>{\hsize=#1\hsize\raggedleft\arraybackslash}X}%
\newcolumntype{C}[1]{>{\hsize=#1\hsize\centering\arraybackslash}X}%
\usepackage{makecell}

\usepackage{url}
\usepackage{algorithm}
\usepackage{algorithmic}
\usepackage{color}
\usepackage{verbatim}
\usepackage{subcaption}
\usepackage[pdftex]{graphicx}
\usepackage{multirow}
\usepackage{fixltx2e}  % for \textsubscript command.
\usepackage{csquotes}
\usepackage{datetime}
\usepackage{tikz}
\usetikzlibrary{arrows}
\usepackage{balance}
\usepackage[cmex10]{amsmath}

\DeclareCaptionLabelSeparator{periodspace}{.\quad}
\def\IEEEbibitemsep{0pt plus .5pt}

\begin{document}

\title{Multimodal Content Analysis for Effective Advertisements on YouTube}

\author{\IEEEauthorblockN{Nikhita Vedula\IEEEauthorrefmark{1},
Wei Sun\IEEEauthorrefmark{1},
Hyunhwan Lee\IEEEauthorrefmark{2}, 
Harsh Gupta\IEEEauthorrefmark{1}, 
Mitsunori Ogihara\IEEEauthorrefmark{3}\IEEEauthorrefmark{4}, 
Joseph Johnson\IEEEauthorrefmark{2}\IEEEauthorrefmark{4}, 
Gang Ren\IEEEauthorrefmark{4}, \\ and
Srinivasan Parthasarathy\IEEEauthorrefmark{1}}
\IEEEauthorblockA{\IEEEauthorrefmark{1}Dept. of Computer Science and Engineering, Ohio State University; \IEEEauthorrefmark{2}Dept. of Marketing, University of Miami;}
\IEEEauthorblockA{\IEEEauthorrefmark{3}Dept. of Computer Science, University of Miami; \IEEEauthorrefmark{4}Center for Computational Science, University of Miami}
Email: \{vedula.5, sun.1868, gupta.749, parthasarathy.2\}@osu.edu, \{aidenhlee, mogihara, jjohnson, gxr467\}@miami.edu}

\maketitle
           
\begin{abstract}

The rapid advances in e-commerce and Web 2.0 technologies have greatly increased the impact of commercial advertisements on the general public. As a key enabling technology, a multitude of recommender systems exists which analyzes user features and browsing patterns to recommend appealing advertisements to users. In this work, we seek to study the characteristics or attributes that characterize an effective advertisement and recommend a useful set of features to aid the designing and production processes of commercial advertisements. We analyze the temporal patterns from multimedia content of advertisement videos including auditory, visual and textual components, and study their individual roles and synergies in the success of an advertisement. The objective of this work is then to measure the effectiveness of an advertisement, and to recommend a useful set of features to advertisement designers to make it more successful and approachable to users. 
Our proposed framework employs the signal processing technique of cross modality feature learning where data streams from different components are employed to train separate neural network models and are then fused together to learn a shared representation. Subsequently, a neural network model trained on this joint feature embedding representation is utilized as a classifier to predict advertisement effectiveness. We validate our approach using subjective ratings from a dedicated user study, the sentiment strength of online viewer comments, and a viewer opinion metric of the ratio of the Likes and Views received by each advertisement from an online platform.   

\end{abstract}

%\begin{IEEEkeywords}
%component; formatting; style; styling
%\end{IEEEkeywords}

\section{Introduction}
\label{sec:introduction}
 
The widespread popularity of the Web and the Internet has led to a growing trend of commercial product publicity online via advertisements. Advertising along with product development, pricing and distribution forms the mix of marketing actions that managers take to sell products and services. It is not enough to merely design, manufacture, price and distribute a product. Managers must communicate, convince and persuade consumers of the competitive superiority of their product for successful sales. This is why firms spend millions of dollars in advertising through media such as TV, radio, print and digital. In 2016, US firms spent approximately \$158 million in advertising. However, despite all this money and effort spent, marketers often find that advertising has little impact on product sales. Effective advertising, defined as advertisements that generate enough sales to cover the costs of advertising, is difficult to create. In fact, John Wanamaker, the originator of the department store concept is reputed to have quipped: ``Half the money I spend on advertising is wasted; the trouble is, I don't know which half.'' Hence, making an effective advertisement that understands its customers' expectations is important for a commercial company. Video advertisements airing on television and social media are a crucial medium of attracting customers towards a product.
 
In a landmark study, Lodish et al.~\cite{lodish1995} examined the sales effects of 389 commercials and found that in a number of cases advertising had no significant impact on sales. There are many reasons that can explain this finding. First, good advertising ideas are rare. Second, people find advertisements annoying and avoid them. Typically, commercials occur within the context of a program that viewers are watching. Therefore, they find the advertisement an unwelcome interruption. Very often we zap out advertisements when we watch TV replays or skip them when it interferes with the digital content we are enjoying. Finally, even when an advertisement manages to hold a viewer’s interest the advertisement may not work because viewers may not pay close enough attention to the message embedded in the advertisement. All these factors make designing advertisement content very challenging and critical to advertising effectiveness.
 
A clear knowledge of the requirements and interests of the specific target group of customers for which the advertisement is meant can go a long way in improving customer satisfaction and loyalty, feedback rate, online sales and the company's reputation. Statistical and knowledge discovery techniques are often used to help companies understand which characteristics or attributes of advertisements contribute to their effectiveness. Apart from product-specific attributes, it is crucial for such techniques to involve a combination of customer-oriented strategies and advertisement-oriented strategies. Many ideas of how to create effective advertisements come from the psychology literature~\cite{cacioppo1984,deighton1989}. Psychologists show that the positive or negative framing of an advertisement, the mix of reason and emotion, the synergy between the music and the type of message being delivered, the frequency of brand mentions, and the popularity of the endorser seen in the advertisement, all go into making an effective advertisement. Another area from which advertisers draw is drama. Thus, the use of dramatic elements such as narrative structure, the cultural fit between advertisement content and the audience are important in creating effective advertisements. But how these ingredients are mixed to develop effective advertisements still remains a heuristic process with different agencies developing their own tacit rules for effective advertising.
 
There are advertisement-specific and user/viewer specific features that can play a role in the advertisement's success. Advertisement-specific features include the context or topic the advertisement is based on, language style or emotion expressed in the advertisement, and the presence of celebrities to name a few. User or viewer specific features include a user's inherent bias towards a particular product or brand, the emotion aroused in the user as a result of watching the advertisement, and user demographics. Many times, users also provide explicit online relevance feedback in the form of likes, dislikes, and comments. These features play an important role in determining the success of an advertisement. Another way advertising agencies improve the chances of coming up with effective advertisements is to create multiple versions of an advertisement and then test it experimentally using a small group of people who represent the target consumer. The hope is that this group of participants will pick the one version of the advertisement that will be effective in the marketplace. The problem with this approach is the production cost of multiple advertisements and the over reliance on the preferences of a small group of participants.
 
The availability of large repositories of digital commercials, the advances made in neural networks and the user generated feedback loop, such as comments likes and dislikes provide us a new way to examine what makes effective advertising. In this paper, we propose a neural-network based approach to achieve our goal, on digital commercial videos. Each advertisement clip is divided into a sequence of frames, from which we extract multimedia visual and auditory features. Apart from these, we also create word-vector embeddings based on the text transcriptions of the online advertisements, which provide textual input to our model. These independent modality features are trained individually on neural networks, to produce high level embeddings in their respective feature spaces. We then fuse the trained models to learn a multimodal joint embedding for each advertisement. This is fed to a binary classifier which predicts whether an advertisement is effective/successful, or ineffective/unsuccessful, according to various metrics of success. We also analyze how the above identified features combine and play a role in making effective and appealing commercials, including the effect of the emotion expressed in the advertisement on the viewer response it garners. {\it The novel methodological contributions of this work lie in the feature engineering and neural network structure design. 
The primary, applied contributions of this work shed light on some key questions governing what makes for a good advertisement and draws insights from the domains of social psychology, marketing, advertising, and finance.}

\section{Related Work}
\label{sec:background}

Previous work has been done in targeted advertisement recommendation to Internet and TV users by exploiting content and social relevance~\cite{wang2013exploiting,yang2007online}. In these works, the authors have used the visual and textual content of an advertisement along with user profile behavior and click-through data to recommend advertisements to users. Content-based multimedia feature analysis is an important aspect in the design and production of multimedia content~\cite{elin2003, cury2012}. Multimedia features and their temporal patterns are known to show high-level patterns that mimic human media cognition and are thus useful for applications that require in-depth media understanding such as computer-aided content creation~\cite{tanahashi2012} and multimedia information retrieval~\cite{lew2006}. The use of temporal features for this is prevalent in media creation and scholarly studies~\cite{campbell2008,huron2008}, movies research~\cite{block2008,bignell2012}, music~\cite{graakjaer2014,cook1992}, and literature~\cite{moretti2007,archer2016}; and these temporal patterns show more ``human-level'' meanings than the plain descriptive statistics of the feature descriptors in these fields. As simple temporal shapes are easy to recognize and memorize, composers, music theorists, musicologists, digital humanists, and advertising agencies utilize them extensively. The studies in~\cite{elin2003,campbell2008} use manual inspection to find patterns, where human analysts inspect the feature visualizations, elicit recurring patterns, and present them conceptually. This manual approach is inefficient when dealing with large multimedia feature datasets and/or where patterns may be across multiple feature dimensions, e.g., the correlation patterns between the audio and the video feature dimensions or between multiple time resolutions. 

We use RNNs and LSTMs in this work to model varied input modalities due to their increased success in various machine learning tasks involving sequential data. CNN-RNNs have been used to generate a vector representation for videos and ``decode'' it using an LSTM sequence model~\cite{yao2015}, and Sutskever et al. use a similar approach in the task of machine translation~\cite{Sutskever2014}. Venugopalan et al.~\cite{venugopalan2016} use an LSTM model to fuse video and text data from a natural language corpus to generate text descriptions for videos. LSTMs have also been successfully used to model acoustic and phoneme sequences~\cite{graves2013,boulanger2013}. Chung et al.~\cite{chung2014} empirically evaluated LSTMs for audio signal modelling tasks. Further, LSTMs have proven to be effective language models~\cite{sundermeyer2012,gulcehre2015}.  

Previous work has focused on modeling multimodal input data with Deep Boltzmann Machines (DBM) in various fields such as speech and language processing, image processing and medical research~\cite{salakhutdinov2009deep,ngiam2011multimodal,srivastava2012multimodal,cao2014,han2015}. In~\cite{salakhutdinov2009deep}, the authors provide a new learning algorithm to produce a good generative model using a DBM, even though the distributions are in the exponential family, and this learning algorithm can support real-valued, tabular and count data. Ngiam et al. in~\cite{ngiam2011multimodal} use DBMs to learn a joint representations of varied modalities. They build a classifier trained on input data of one modality and test it on data of a different modality. In~\cite{srivastava2012multimodal}, the authors build a multimodal DBM to infer a textual description for an image based on image-specific input features, or vice versa.

%Another work is done using multimodal fusion~\cite{yang2007online} where even the user's profile was not needed. The textual, visual, and aural relevance of videos were extracted and relevance feedback is used to automatically adjust intra-weights within each modality and inter-weights among different modalities by users' click-though data, as well as attention fusion function to fuse multimodal relevance together. 

Lexical resources such as WordNetAffect~\cite{strapparava2004wordnet}, SentiWordNet~\cite{esuli2007sentiwordnet} and the SentiFul database~\cite{neviarouskaya2011sentiful} have long been used for emotion and opinion analysis. Emotion detection has been done using such affective lexicons with distinctions based on keywords and linguistic rules to handle affect expressed by interrelated words~\cite{bao2012mining,rao2014building,strapparava2007semeval}. The first work on social emotion classification was the SWAT algorithm from the SemEval-2007 task~\cite{strapparava2007semeval}. In~\cite{bao2012mining}, the authors propose a emotion detection model based on the Latent Dirichlet Allocation. This model can leverage the terms and emotions through the topics of the text. In~\cite{rao2014building}, the authors propose two kinds of emotional dictionaries, word-level and topic level, to detect social emotions. In recent years, CNNs and RNNs have been utilized to effectively perform emotion detection~\cite{rao2014,wollmer2013}.

\section{Methodology}
\label{sec:methodology}

In this section, we begin with a description of the multimedia temporal features we have extracted and employed, based on video frames, audio segments and textual content of commercial advertisements; followed by creating a joint embedding of these multimodal inputs. We then describe our method to detect emotion in the advertisements' linguistic content.
%{\bf explain why emotion detection is done??}

\smallskip
\subsection{Feature Extraction}
\label{sec:featureextraction}

\smallskip
\subsubsection{Visual (video) Features}
\label{sec:videofeatures}
\begin{figure}
\centering
    \includegraphics[width=\linewidth,height=6cm,keepaspectratio]{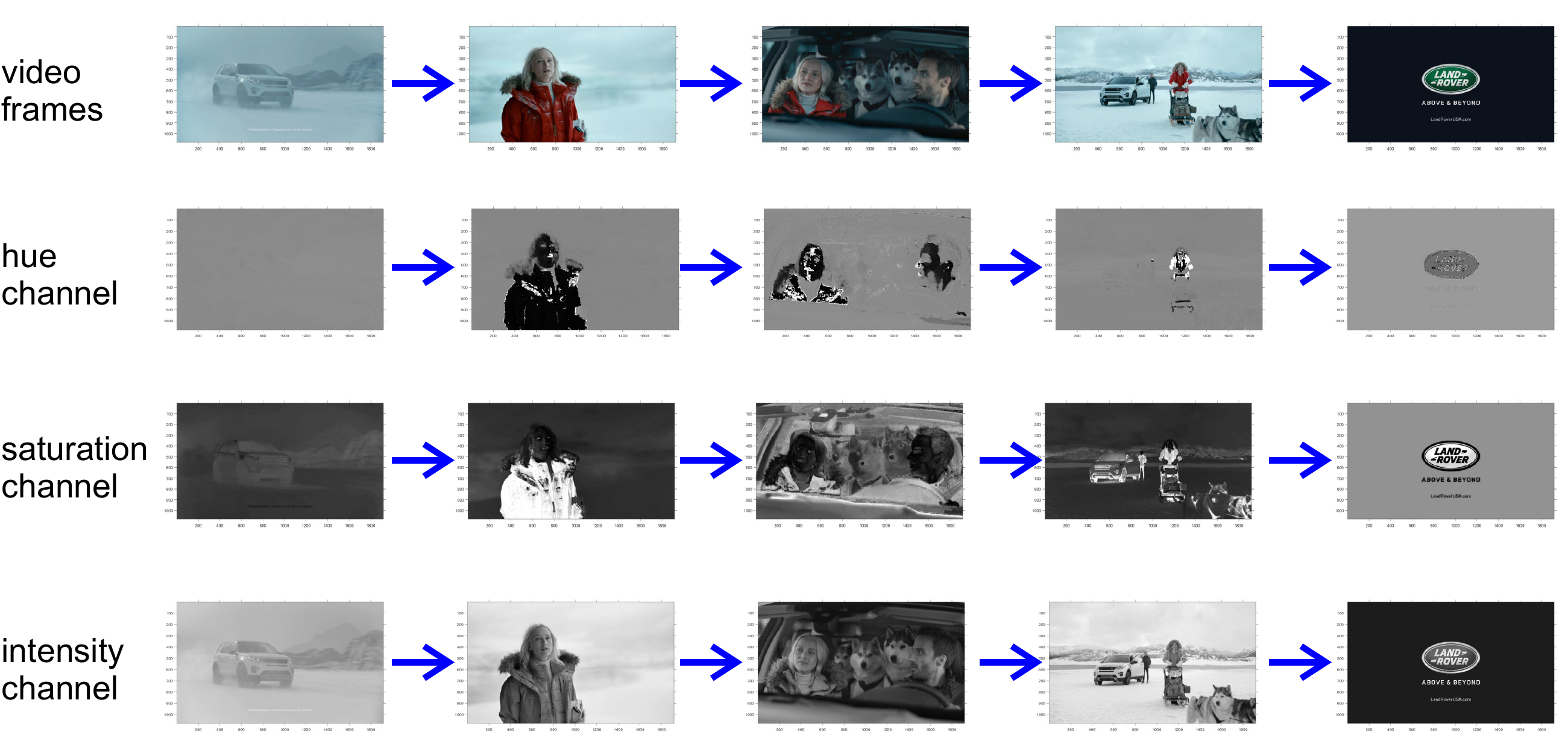}
    \caption{Multimedia timeline analysis of three video signal dimensions.}
    \label{fig:videofeat}
\end{figure}

The video features of content timelines are extracted from the image features from sampled video frames. For speeding up the signal processing algorithms, one in ten video frames is sampled and measured for video feature extraction. For each pixel in a sampled video, we measure the hue, saturation and brightness values as in~\cite{goldstein2016}. The hue dimension reflects the dominant color or its distribution and is one of the most important post-production and rendering decisions~\cite{bignell2012}. The saturation dimension measures the extent to which the color is applied, from gray scale to full color. The brightness dimension measures the intensity of light emitted from the pixel. These three feature dimensions are closely related to human perception of color relationships~\cite{bignell2012}, so this measurement process serves as a crude model of human visual perception (Figure~\ref{fig:videofeat}).

The feature descriptors for each video frame include the mean value and spatial distribution descriptors of the hue-saturation-brightness values of the constituent pixels. For measuring the deviations of these feature variables at different segments of the screen, the mean values of the screen's sub-segments and the differences between adjacent screen segments are calculated. The above video features are mapped to their time locations to form high-resolution timelines. We also segment the entire time duration of each video into 50/20/5 time segments as a hierarchical signal feature integration process and calculate the temporal statistics inside each segment including temporal mean and standard deviation, as well as the aggregated differences between adjacent frames. 

\smallskip
\subsubsection{Auditory (audio) Features}
The audio signal features include auditory loudness, onset density, and timbre centroid. Loudness is based on a computational auditory model applied on the frequency-domain energy distribution of short audio segments~\cite{muller2015}. We first segment the audio signal into 100 ms short segments ensuring enough resolution in time and frequency domains, calculate the fast Fourier transform for each, and utilize the spectral magnitude as the frequency-energy descriptor. Because the human auditory system sensitivity varies with frequency, a computational auditory model~\cite{moore2012} is employed to weight the response level to the energy distribution of audio segments. The loudness $L_a$ is thus calculated as: 

\begin{center}
$L_a = log_{10} \sum_{k=1}^K S(k)\eta(k) $
\end{center}

where $S(k)$ and $\eta(k)$ denote the spectral magnitude and frequency response strength respectively at frequency index $k$. $K$ is the range of the frequency component. Similar to the temporal resolution conversion algorithm in Section~\ref{sec:videofeatures}, the loudness feature sequence is segmented and temporal characteristics like the mean and standard deviation in each segment are used as feature variables. 

For high resolution tracks, the audio onset density measures the time density of sonic events in each segment $1/50^{th}$ of the entire video duration (2 s). The onset detection algorithm~\cite{muller2015} records onsets as time locations of large spectral content changes, and the amount of change as the onset significance. For each segment, we count onsets with significance value higher than a threshold and normalize it by the segment length as the onset density. We use longer segments because of increased robustness in onset detection. For the same reason, the onset density of lower time resolutions is measured from longer segments $1/20^{th}$ or $1/5^{th}$ of the total length, and not from the temporal summarization of the corresponding high resolution track. The timbre dimensions are measured from short 100 ms segments, similar to loudness. The timbre centroid $T_c$ is measured as: 
 
\begin{center}
$T_c = \frac{\sum_{k=1}^K k S(k)}{\sum_{k=1}^K S(k)}$
\end{center}

The hierarchical resolution timbre tracks are summarized in a similar manner as auditory loudness. 

\smallskip
\subsubsection{Textual Features}
Word2vec~\cite{mikolov2013} is a successful approach to word vector embedding, which uses a two-layer neural network with raw text as an input to generate a vector embedding for each word in the corpus. After preliminary experiments with some other word embedding strategies~\cite{pennington2014glove,levy2014neural}, we decided on word2vec since we found its embeddings to be more suitable for our purpose. We first pre-processed and extracted the text transcription of each advertisement to get a list of word tokens. We then used the 300-dimensional word vectors pre-trained on the Google News Dataset, from \url{https://code.google.com/archive/p/word2vec/} to obtain a word embedding for each token.  

\subsection{Learning Multimodal Feature Representations}

\smallskip
\subsubsection{LSTMs for Sequential Feature Description}

\begin{figure}
\centering
    \includegraphics[width=\linewidth,height=4.5cm,keepaspectratio]{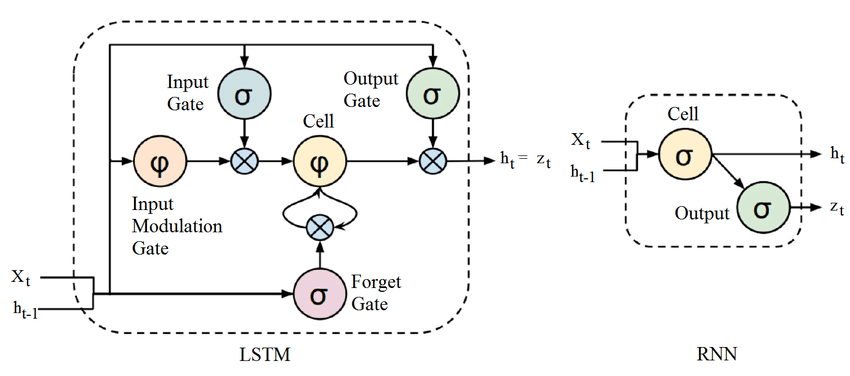}
    \caption{LSTM cell unit as described in~\cite{fayyaz2016}, showing the three sigmoidal gates and the memory cell.}
    \label{fig:lstmcell}
\end{figure}

A Recurrent Neural Network (RNN) generalizes feed forward neural networks to sequences, that is, they learn to map a sequence of inputs to a sequence of outputs, for which the alignment of inputs to the outputs is known ahead of time~\cite{Sutskever2014}. However, it is challenging to use RNNs to learn long-range time dependencies, which is handled quite well by LSTMs~\cite{hochreiter1997}. At the core of the LSTM unit is a memory cell controlled by three sigmoidal gates, at which the values obtained are either retained (when the sigmoid function evaluates to 1) or discarded (when the sigmoid function evaluates to 0). The gates that make up the LSTM unit are: the input gate $i$ deciding whether the LSTM retains its current input $x_t$, the forget gate $f$ that enables the LSTM to forget its previous memory context $c_{t-1}$, and the output gate $o$ that controls the amount of memory context transferred to the hidden state $h_t$. The memory cell thus can encode the knowledge of inputs observed till that time step. The recurrences for the LSTM are defined as:

\begin{center}
$i_t = \sigma(W_{xi}x_t + W_{hi}h_{t-1})$ \\
$f_t = \sigma(W_{xf}x_t + W_{hf}h_{t-1})$ \\
$o_t = \sigma(W_{xo}x_t + W_{ho}h_{t-1})$ \\
$c_t = f_t \odot c_{t-1} + i_t \phi(W_{xc}x_t + W_{hc}h_{t-1})$ \\
$h_t = o_t \odot \phi(c_t)$
\end{center}

where $\sigma$ is the sigmoid function, $\phi$ is the hyperbolic tangent function, $\odot$ represents the product with the gate value and $W_{ij}$ are the weight matrices consisting of the trained parameters.

\begin{figure}
\centering
    \includegraphics[width=\linewidth,height=4.5cm,keepaspectratio]{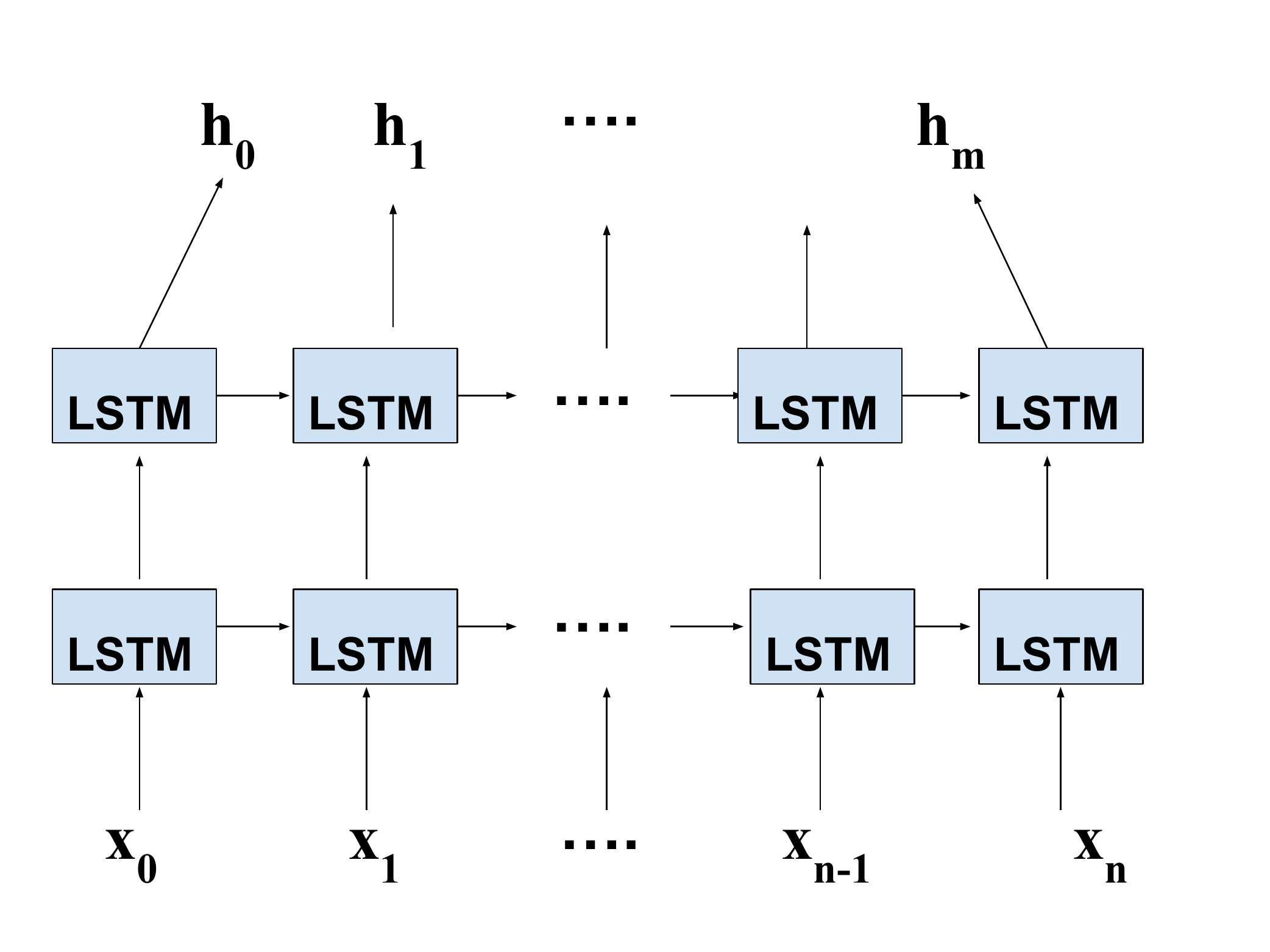}
    \caption{LSTM model with two hidden layers, each layer having 100 hidden units each, used for training individual input modalities.}
    \label{fig:lstmmodel}
\end{figure}

We use an LSTM model with two layers to encode sequential multimedia features, employing a model of similar architecture for all the three input modalities. Based on the features described in Section~\ref{sec:featureextraction}, we generate a visual feature vector for temporal video frames of each advertisement, which forms the input to the first LSTM layer of the video model. We stack another LSTM hidden layer on top of this, as shown in Figure~\ref{fig:lstmmodel}, which takes as input the hidden state encoding output from the first LSTM layer. Thus, the first hidden layer would create an aggregated encoding of the sequence of frames for each video, and the second hidden layer encodes the frame information to generate an aggregated embedding of the entire video.

We next generate an audio feature vector for the temporal audio segments described in Section~\ref{sec:featureextraction}, and encode it via a two hidden layer LSTM model. Finally, for the textual features, we first encode the 300-dimensional word vector embedding of each word in the advertisement text transcription through the first hidden layer of an LSTM model. A second LSTM hidden layer is applied to this encoding to generate an output summarized textual embedding for each advertisement.

\smallskip
\subsubsection{Multimodal Deep Boltzmann Machine (MDBM)}

A classical Restricted Boltzmann Machine (RBM)~\cite{smolensky1986information} is an undirected graphical model with binary-valued visible layers and hidden layers~\cite{hinton2006reducing,freund1994unsupervised}. We use the Gaussian-Bernoulli variant of an RBM which can model real-valued input data, vertically stacking the RBMs to form a Deep Boltzmann Machine (DBM)~\cite{srivastava2012multimodal, salakhutdinov2009deep}. We use three DBMs to individually model the visual, auditory and textual features. Each DBM has one visible layer $\mathbf{v}\in R^{n}$, where n is the number of visible units, and two hidden layers $\mathbf{h_{i}}\in\left\{0,1\right\}^{m}$, where m is the number of hidden units and $i=1,2$. 

A DBM is an energy based generative model. Therefore, the energy of the joint state $\left\{\mathbf{v}, \mathbf{h}^{\left(1\right)}, \mathbf{h}^{\left(2\right)}\right\}$ can be defined as follows:

\begin{gather*}
\nonumber
\label{DBM_visible_dis}
P\left ( \mathbf{v}; \theta \right ) = \sum_{\mathbf{h^{(1)}}, \mathbf{h^{(2)}}} P\left ( \mathbf{v}, \mathbf{h^{(1)}}, \mathbf{h^{(2)}};\theta \right ) \\
= \frac{1}{Z\left ( \theta \right )}\sum_{\mathbf{h^{(1)}}, \mathbf{h^{(2)}}} exp\left ( -E\left ( \mathbf{v}, \mathbf{h}; \theta \right ) \right)
\end{gather*}  

\begin{gather*}
\nonumber
\label{DBM_visible_dis_2}
E\left ( \mathbf{v}, \mathbf{h}; \theta \right ) = \sum_{i} \frac{\left ( v_{i}-b_{i} \right )^{2}}{2\sigma _{i}^{2}} - \sum_{ij}\frac{v_{i}}{\sigma _{i}}W_{ij}^{(1)}h_{j}^{(1)} \\
- \sum_{jk}W_{jk}^{(2)}h_{j}^{(1)}h_{k}^{(2)}
 -\sum_{j}b_{j}^{(1)}h_{j}^{(1)} - \sum_{k}b_{k}^{(2)}h_{k}^{(2)}
\end{gather*}  

\noindent where $\mathbf{h} = \left \{ \mathbf{h}^{(1)}, \mathbf{h}^{(2)} \right\}$ denotes the units of two hidden layers and $\theta = \left \{ \mathbf{W^{^{(1)}}}, \mathbf{W^{^{(2)}}}, \mathbf{b^{(1)}}, \mathbf{b^{(2)}} \right \}$ denotes the weights and bias parameters of the DBM model. 

\noindent $Z\left ( \theta \right ) = \int_\mathbf{v} \sum _\mathbf{h} exp\left ( -E\left ( \mathbf{v}, \mathbf{h}; \theta \right ) \right )dv$ denotes the partition function. 

\begin{figure}
\centering
    \includegraphics[width=\linewidth,height=4.5cm,keepaspectratio]{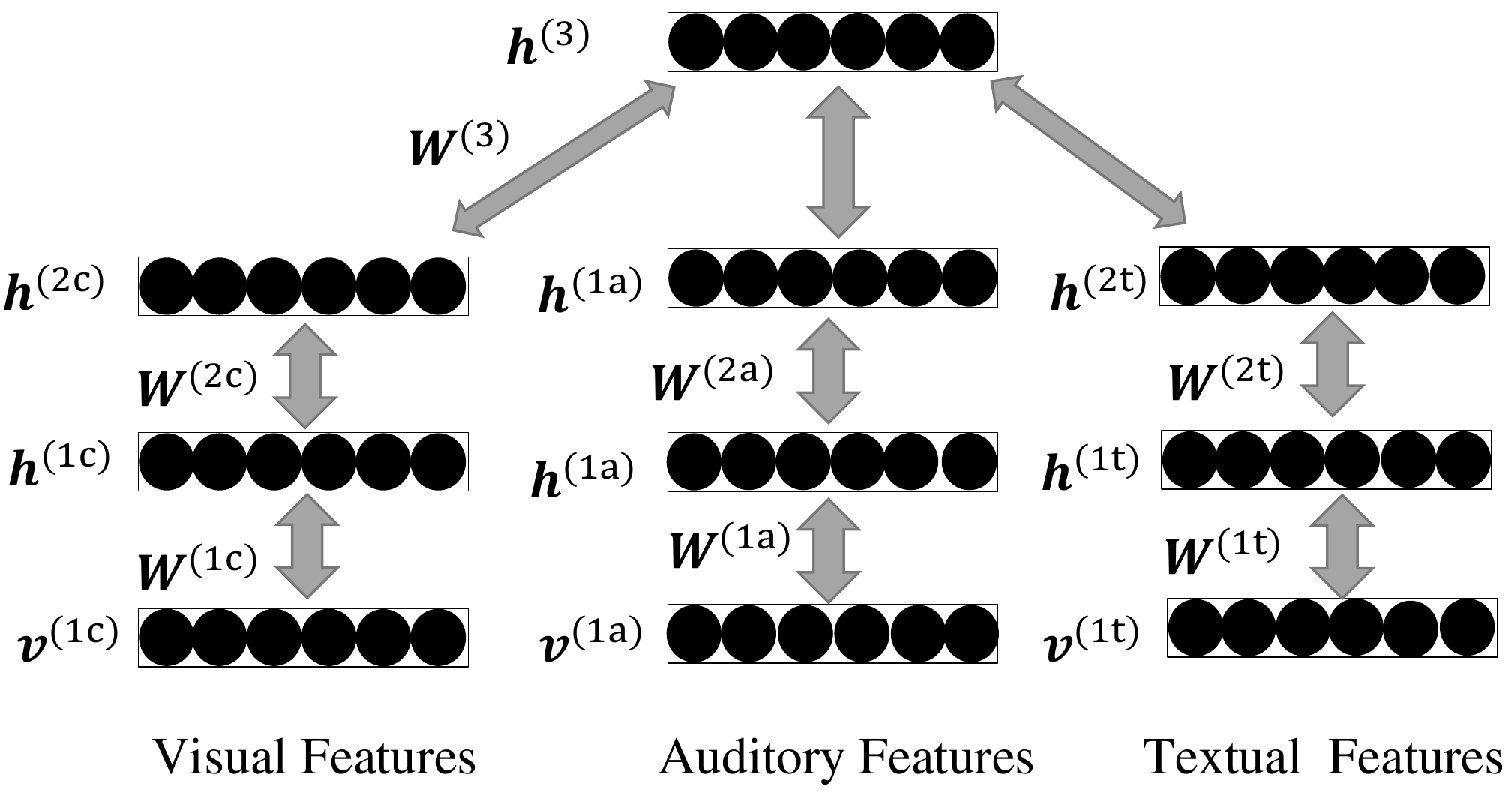}
    \caption{MDBM that models the joint distribution over the visual features, auditory features and textual features. All layers in this model are binary layers except for the bottom real valued layer.}
    \label{fig:mdbm}
\end{figure}

We formulate a multimodal DBM~\cite{srivastava2012multimodal} by combining the three DBMs and adding one additional layer at the top of them, as in Figure~\ref{fig:mdbm}. The joint distribution over the three kinds of input data is thus defined as:

\begin{center}
\label{MDBM_distribution}
$P\left ( \mathbf{v}^{c}, \mathbf{v}^{a}, \mathbf{v}^{t}; \theta \right ) = \frac{1}{Z\left ( \theta \right )} 
\sum_{\mathbf{h}}exp\left ( -V - A - T + J\right )$
\end{center}
where $V$, $A$ and $T$ represent the visual, auditory and textual pathways respectively, and $J$ represents the joint layer at the top.

\begin{gather*}
V = \sum_{i}\frac{(v_{i}^{c}-b_{i}^{c})^{2}}{2\sigma_{i}^{2}} - \sum_{ij}\frac{v_{i}^{c}}{\sigma_{i}}W_{ij}^{(1c)}h_{j}^{(1c)} -\sum_{jl}W_{jl}^{(2c)}h_{j}^{(1c)}h_{l}^{(2c)} \\ -  \sum_{j}b_{j}^{(1c)}h_{j}^{(1c)} - \sum_{l}b_{l}^{(2c)}h_{l}^{(2c)}; \hspace{0.2cm} J= \\
\vspace{0.4cm}
\sum_{lp}W^{(3c)}h_{l}^{(2c)}h_{p}^{(3)} + \sum_{lp}W^{(3a)}h_{l}^{(2a)}h_{p}^{(3)} + \sum_{lp}W^{(3t)}h_{l}^{(2t)}h_{p}^{(3)}. 
\end{gather*}

$A$ and $T$ have similar expressions as $V$. 

$\mathbf{v^{c}}$, $\mathbf{v^{a}}$ and $\mathbf{v^{t}}$ denote the visual, auditory and textual feature inputs over their respective pathways of $V$, $A$ and $T$. $\mathbf{h} = \left \{ \mathbf{h^{\left (1c  \right )}}, \mathbf{h^{\left (2c  \right )}}, \mathbf{h^{\left (1a  \right )}},\mathbf{h^{\left (2a  \right )}}, \mathbf{h^{\left (1t  \right )}}, \mathbf{h^{\left (2t  \right )}},  \mathbf{h^{\left (3  \right )}}\right \}$ denotes the hidden variables, $\mathbf{W}$ denotes the weight parameters, and $\mathbf{b}$ denotes the biases.

We first pre-train each modality-specific DBM individually with greedy layer-wise pretraining~\cite{hinton2006reducing}. Then we combine them together and regard it as a Multi Layer Perceptron~\cite{ruck1990multilayer} to tune the parameters that we want to learn.

% where the Visual Feature Pathway V is : 
% \begin{equation}
% \nonumber
% \label{MDBM_video}
% %\begin{split}
% -\sum_{i}\frac{(v_{i}^{c})^{2}}{2\sigma_{i}^{2}} + \sum_{ij}\frac{v_{i}^{c}}{\sigma_{i}}W_{ij}^{(1c)}h_{ij}^{(1c)} +\sum_{jl}W_{jl}^{(2c)}h_{j}^{(1c)}h_{l}^{(2c)} \\ +  \sum_{j}b_{j}^{(1c)}h_{j}^{(1c)} + \sum_{l}b_{l}^{(2c)}h_{l}^{(2c)}
% %\end{split}
% \end{equation}
% the Audio Feature Pathway A is:
% \begin{equation}
% \nonumber
% \label{MDBM_audio}
% %\begin{split}
% -\sum_{i}\frac{(v_{i}^{a})^{2}}{2\sigma_{i}^{2}} + \sum_{ij}\frac{v_{i}^{a}}{\sigma_{i}}W_{ij}^{(1a)}h_{ij}^{(1a)} +   \sum_{jl}W_{jl}^{(2a)}h_{j}^{(1a)}h_{l}^{(2a)} \\ + \sum_{j}b_{j}^{(1a)}h_{j}^{(1a)} + \sum_{l}b_{l}^{(2a)}h_{l}^{(2a)}
% %\end{split}
% \end{equation}
% the Textual Feature Pathway T is:
% \begin{equation}
% \nonumber
% \label{MDBM_text}
% %\begin{split}
% -\sum_{i}\frac{(v_{i}^{t})^{2}}{2\sigma_{i}^{2}} + \sum_{ij}\frac{v_{i}^{t}}{\sigma_{i}}W_{ij}^{(1t)}h_{ij}^{(1t)} + \sum_{jl}W_{jl}^{(2t)}h_{j}^{(1t)}h_{l}^{(2t)} \\ + \sum_{j}b_{j}^{(1t)}h_{j}^{(1t)} + \sum_{l}b_{l}^{(2t)}h_{l}^{(2t)} 
% %\end{split}
% \end{equation}
% and the Joint Layer J is:
% \begin{equation}
% \nonumber
% \label{MDBM_joint}
% %\begin{split}
% \sum_{lp}W^{(3c)}h_{l}^{(2c)}h_{p}^{(3)} + \sum_{lp}W^{(3)}h_{l}^{(2a)}h_{p}^{(3)} + \sum_{lp}W^{(3t)}h_{l}^{(2t)}h_{p}^{(3)}
% %\end{split}
% \end{equation}

\begin{comment}
\begin{figure}
\centering
\begin{minipage}{.5\textwidth}
  \centering
  \includegraphics[width=.7\linewidth,height=5cm,keepaspectratio]{Figures/lstmmodel}
  \caption{LSTM model with two hidden \\ layers, each layer having 100 hidden units, \\ for training individual input modalities.}
  \label{fig:lstmmodel}
\end{minipage}%
\begin{minipage}{.5\textwidth}
  \centering
  \includegraphics[width=.98\linewidth,height=13cm,keepaspectratio]{Figures/MDBM}
  \caption{MDBM modeling the joint distribution over the three modalities. All layers in this model are binary except for the bottom real valued layer.}
  \label{fig:mdbm}
\end{minipage}
\end{figure}
\end{comment}

\smallskip
\subsubsection{Inferring a Joint Representation}

\begin{figure}
\centering
    \includegraphics[width=\linewidth,height=4.5cm,keepaspectratio]{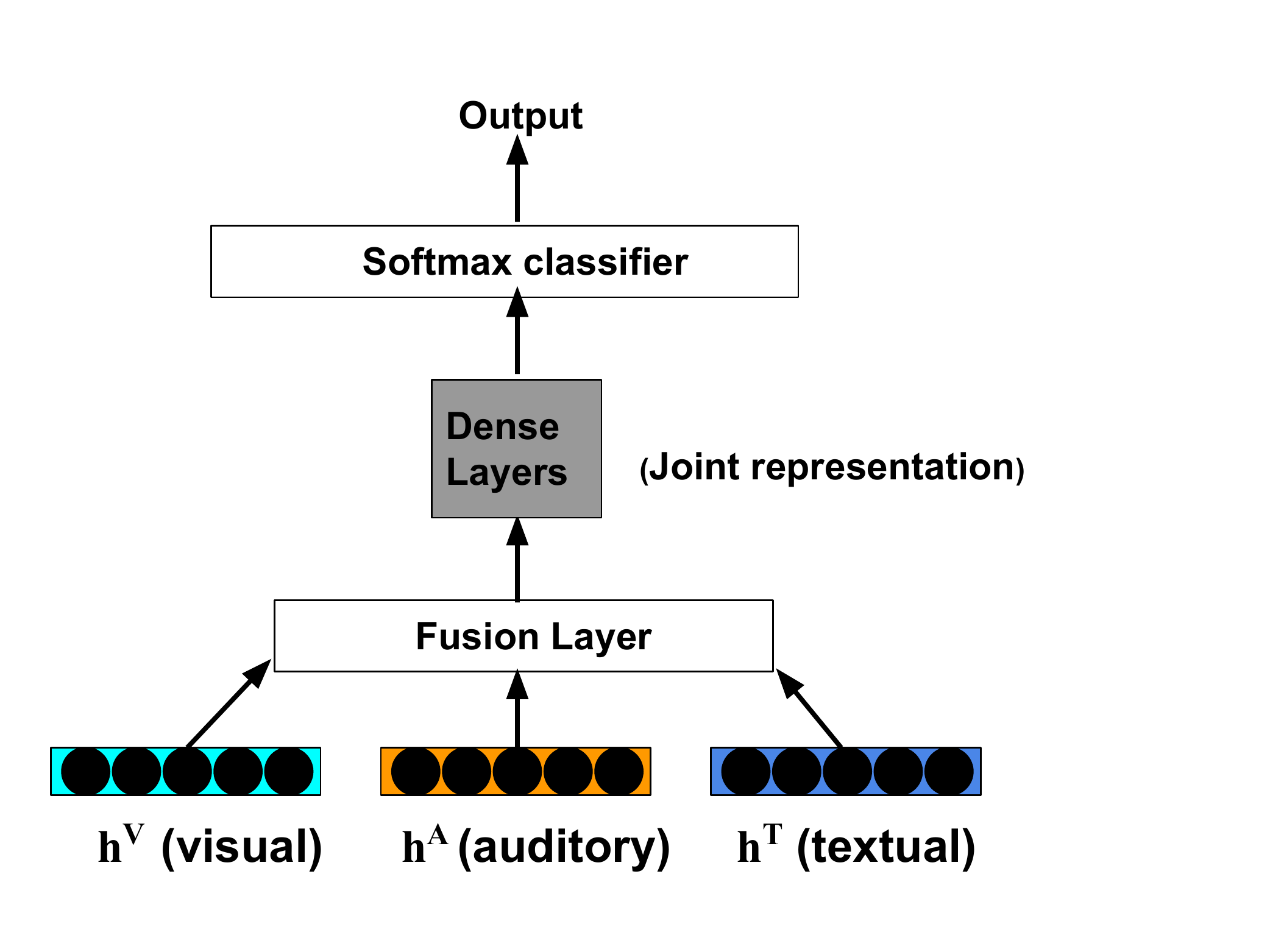}
    \caption{Multimodal LSTM/DBM model that learns a joint representation over visual, auditory and textual features, followed by a softmax classifier.}
    \label{fig:jointmodel}
\end{figure}

In order to avoid our learning algorithm getting stuck on local optima, we normalized the visual, auditory and textual input data into a uniform distribution. Once we obtain high-level feature embeddings ($\mathbf h^{V}, \mathbf h^{A}, \mathbf h^{T}$) from the final hidden layer of the three respective models of audio, video and text, we concatenate the three hidden layer embeddings in a layer called the {\it fusion layer}, which enables us to explore the correlation between the three kinds of features (see Figure~\ref{fig:jointmodel}). 
In order to minimize the impact of overfitting, we perform dropout regularization~\cite{srivastava2014} on the fusion layer with a dropout probability of 0.5. The combined latent vector is passed through multiple dense layers with non-linear activation functions (we used ReLU), before being passed through a final softmax layer to predict the output class of the advertisement. We assume a binary classifier for the advertisements with two classes: effective or successful, and ineffective or unsuccessful. Thus, the probability of predicting a class label $y$ is:

\begin{center}
$p(y|\mathbf {x_V}, \mathbf {x_A}, \mathbf {x_T}) \propto exp(W[\mathbf {h^{V}}; \mathbf {h^{A}}; \mathbf {h^{T}}] + \mathbf b) $
\end{center}
where $y$ denotes the class, $\mathbf {x_V}, \mathbf {x_A}, \mathbf {x_T}$ are the video, audio and text features of advertisement $x$, $W$ is the weight matrix, [u;v] denotes the concatenation operation and $\mathbf{b}$ the biases.

\subsection{Analyzing Emotional Content}
\label{sec:emotiondetection}

The affective emotions present in the linguistic content or style of an advertisement can play a crucial role in invoking positive feelings towards it in its viewers. Therefore, we detected the dominant emotion reflected in an advertisement's textual content, and compared its correlation with the actual feelings it induced in a sample of viewers (more on this in Section~\ref{sec:experiments}). We used two approaches to detect the dominant emotion prevalent in the linguistic content and style of each advertisement, out of joy, surprise, anticipation, trust, sadness, anger, fear, and disgust. In the first approach, we modeled this as a multi-class classification problem using a simple neural network consisting of a single LSTM layer with 0.6 dropout regularization. We trained this on text from a mixture of the ISEAR dataset~\cite{isear} and the NRC Word-Emotion association lexicon~\cite{mohammad2013}, and 0.6 dropout to input and recurrent connections of the memory units with LSTM. The output was fed into a multi-class softmax classifier to detect the distribution of emotions in the text. We tested this model on the advertisement text transcriptions to generate their emotion distribution.
 
We also experimented with a dictionary-based approach, where we manually constructed a dictionary of words associated with each emotion from sources such as the NRC Emotion Lexicon~\cite{mohammad2013}, ISEAR~\cite{isear} and WordNetAffect~\cite{strapparava2004wordnet}, and computed the dominant emotion for each advertisement based on a text comparison with the dictionary to construct additional features to our network model. Though there is no consideration of semantic relations between words here, surprisingly on manual validation we found its results to be just slightly less effective than the LSTM-based model. Adding these features to the LSTM model did not appear to yield any additional
benefits (suggesting that the LSTM model was automatically learning similar relationships).

\section{Experiments}
\label{sec:experiments}
 
\smallskip
\subsection{Dataset and Evaluation Metrics}
We evaluated our proposed methodology on a dataset of 200 advertisements crawled online from the website {\it YouTube}, spanning different categories such as food and beverages, movie trailers, health and medicine, and clothing. The ground truth for whether an advertisement is successful/effective or not was based on three independent metrics, detailed below.  
 
%\subsection{Evaluation Metrics}
\label{sec:groundtruth}
%The ground truth for whether an advertisement is successful/effective or not was based on three independent metrics described in detail below.
%: an extensive user study done on all advertisements considered; the sentiment strength of the comments which were given by people who watched the video on Youtube and expressed their views on it; and third, on the ratio of the number of likes received by each advertisement video on Youtube to the total number of views or visits it received by users. We describe these in detail below.
 
\smallskip
\subsubsection{User Study}

First, a user study was conducted on the 200 advertisement video clips using the Qualtrics survey platform. The test environment included a professional video playback workstation, a 24-inch color accurate production monitor, a rating laptop, and a high quality loudspeaker system (average sound pressure level = 56 dB SPL). The test room had sound absorption material in one wall and a reverberation time of 0.4 sec RT60). Only one test subject was in the test room during a rating session. The subjects were allowed to watch the videos in any order or re-watch a video anytime. The questionnaire included 96 questions with categories described below:

\begin{enumerate}
\item Affective Response, including adjectives of emotion descriptors such as ``I feel pleasant", ``the main emotion of this video is sadness", and ``the video makes me feel enthusiastic";

\item Attention patterns and message relevance, such as ``the video message is important to me", ``the product information is worth remembering", and ``how much attention did you pay to the video";

\item Attitude toward the advertisement, such as ``the commercial was trustworthy", ``the commercial was conclusive", and ``the commercial is catchy";

\item Attitude toward the brand, such as ``I need lots more information about this brand before I buy it", ``I know what I'm going to get from this brand", and ``I know I can count on this brand being there in the future";

\item Attitude toward the products, such as ``the product can help me satisfy my desires", ``the product is cutting edge technology", and ``the product is durable";

\item Persuasiveness and purchase intensions, such as ``this commercial changed my attitude towards the brand", ``this commercial will influence my shopping habits", and ``would you buy this brand if you happened to see it in a store";

\end{enumerate}

Most questions solicited ratings ranging from `Strongly Disagree' (rated 1) to `Strongly Agree' (rated 7). Questions in category 2 used ratings from `Not At All (paying attention)' (1) to `A lot' (7). Three Questions in category 6 solicited binary answers as `Yes' (1) or `No' (2), averaged over all questions and categories. We considered the advertisements with a mean rating $\leq3$ (averaged over all questions) as ineffective, and the rest as effective advertisements. The rating results were anonymized, and the experiment content and procedures were approved by the  Internal Review Board of the respective organizations involved (IRB number 2015B0249)%\footnote{Types  of organizations involved, IRB Protocol numbers, and participant profile details are omitted due to ICDM triple blind review requirements.}
, and meet with standard Nielson Norman Group guidelines.

%Firstly, a user study was conducted on the 200 advertisements, comprising 96 questions on categories such as brand presence and reliability, emotion expressed in the advertisement, emotion induced in the user, usefulness, attention paid while watching, commercial presence and its influence on the user.
%and there were 900 responses reported based on the study. Each response was based on 96 questions.
%On an average each advertisement was rated by four different users in order to have a well formed response and avoid any bias towards brand or the product. %The 96 questions were based on 11 broad categories where each category had different number of questions on it.Ratings ranged from `Strongly Agree'(+3) to `Strongly Disagree'(-3) and were mapped to numerical values for the purpose of evaluation. `Strongly agree' was given a rating of positive 3, `agree' was given a rating of 2, `somewhat agree' 1 and an overall rating was calculated according to the response in the user study.
%The rationale behind this approach is that if any advertisement's rating is greater than the base minimum liking of an advertisement by any user rating then we can certainly say that it is an effective advertisement.
 
\smallskip
\subsubsection{Sentiment Strength}
Second, we scraped the comments expressed by users on YouTube on each advertisement, and calculated the strength of the sentiment expressed in them using a tool called SentiStrength~\cite{sentistrength}. The sentiment strength scores ranged from -5 to 5, and all advertisements having a mean score above a threshold of $2.5$ were considered as effective, and the rest as ineffective advertisements.
 
\smallskip
\subsubsection{Likes per Visits}
Third, the number of `likes' i.e. explicit positive feedback or appreciation received by an advertisement video on YouTube is a clear indicator of its popularity among its viewers. We used the measure of the likes an advertisement receives over the total number of times it was viewed or visited as a measure of its effectiveness. YouTube also provides us with the number of `dislikes' or negative feedback received by the advertisement, but since this number was always lesser than the number of likes the advertisements received, we did not consider this quantity in our assessment. All advertisements having a likes to views ratio above the mean of the ratio values received by all advertisements were considered as effective, and the rest as ineffective advertisements.
 
\smallskip
\subsection{Results}
 
\smallskip
\subsubsection{Classification}
 
%Baselines
We compare our method against the baseline classifiers of Linear SVM and Logistic Regression, which take as input a concatenation of the visual, auditory, and textual features. We trained our neural network models over 15 epochs, minimizing the binary cross entropy loss function using the Adam~\cite{kingma2014adam} optimizer, with an initial learning rate of 0.001. We employed 150 randomly selected advertisements for training and 50 for testing our method, and averaged our results over 50 runs.
 
\begin{table}[!t]
\caption{Classification results using various classifiers and ground truth metrics (best performance in bold)}
%\vspace{-0.6em}
\label{tab:result}
\small
\begin{tabularx}{\linewidth}{L{0.74}L{0.78}L{0.32}L{0.16}}
\hline
{\bf Model} & {\bf Ground truth} & {\bf Accuracy} & {\bf F1} \\
\hline
Linear SVM & Comment sentiment & 0.58 & 0.565 \\
%\hline
Linear SVM & Likes/visits & 0.586 & 0.568 \\
%\hline
Linear SVM & User study & 0.565 & 0.541 \\
%\hline
Logistic Regression & Comment sentiment & 0.468 & 0.44 \\
%\hline
Logistic Regression & Likes/visits & 0.55 & 0.529 \\
%\hline
Logistic Regression & User study & 0.542 & 0.52 \\
%\hline
Multimodal DBM & Comment sentiment & 0.60  & 0.66  \\
%\hline
Multimodal DBM & Likes/visits & 0.61 & 0.71 \\
%\hline
Multimodal DBM & User study & 0.66 & 0.64 \\
%\hline
Multimodal LSTM & Comment Sentiment & 0.786 & 0.765 \\
%\hline
Multimodal LSTM & Likes/visits & 0.8 & 0.769 \\
%\hline
Multimodal LSTM & User study & {\bf 0.83} & {\bf 0.81}\\
%\hline
Video-only LSTM & Comment Sentiment & 0.34 & 0.334 \\
%\hline
Video-only LSTM & Likes/visits & 0.39 & 0.378 \\
%\hline
Video-only LSTM & User study & 0.44 & 0.408\\
%\hline
Audio-only LSTM & Comment sentiment & 0.365 & 0.341 \\
%\hline
Audio-only LSTM & Likes/visits & 0.401 & 0.4 \\
%\hline
Audio-only LSTM & User study & 0.416 & 0.37 \\
%\hline
Text-only LSTM & Comment sentiment & 0.478 & 0.445 \\
%\hline
Text-only LSTM & Likes/visits & 0.49 & 0.468 \\
%\hline
Text-only LSTM & User study & 0.52 & 0.52 \\
\hline
\end{tabularx}
\end{table}
 
Table~\ref{tab:result} displays the F1-score and accuracy of various classifiers. We find that {\it the multimodal LSTM model is able to achieve the best accuracy and F1-score of $>0.8$ as compared to other models, and the difference in accuracy is significant}. The last three rows of the table represent the performance by using individual LSTM models with a single input modality. We find that using a multimodal joint feature representation gives us a huge advantage over any of the individual models. The LSTM model that classifies advertisements based only on textual word2vec features appears to perform the best as compared to the other two models (video-only and audio-only, whose accuracies are below 50\%).
 
Drilling down into the quality of our LSTM-based classifier, we found that it had a False Positive Rate of 0, i.e. it did not misclassify any seemingly ineffective advertisements as effective. All its misclassifications were False Negatives, reporting effective advertisements as ineffective. In addition, in order to study the effect of the presence of the name of a brand on the advertisement success, we investigated the performance of our model by removing all occurrences of brand name from the advertisement text. We found the accuracy of the Text-only model to reduce to nearly 46\%, while the accuracy of the multimodal LSTM dropped down to about 73\%. This confirms that the presence of brand name does play an important role in determining the success of an advertisement. We also inspected the impact of the position of the brand name in the advertisement text i.e. its occurrence in the beginning, middle or end of the advertisement. We did not find any significant difference in the performance of the Text-only LSTM or the joint LSTM model.
 
\smallskip
\subsubsection{Comparing Ground Truth Measures and Model Selection}
 
As displayed in Table~\ref{tab:result}, we evaluate the performance of four different algorithms using the three ground truth measures
%. The three ground truth measures are as follows and have been
described in Section~\ref{sec:groundtruth}. %earlier: an extensive user study done on all advertisements considered, the sentiment strength of the comments posted by people who watched the advertisement on Youtube, and the ratio of the number of likes received by each advertisement video on Youtube to the total number of views or visits it received from the public.
Information about the views and likes received by an advertisement online is the easiest to obtain and directly provides a real-time opinion of the general public on its success. However, with this measure it is often difficult to control for adversarial effects and noise (e.g. arbitrarily liking or disliking a video), and it lacks provenance. The metric using comment sentiment strength is also easy to acquire and compute, however its quality and efficacy might be somewhat limited by the accuracy of the sentiment detection algorithm in addition to the other factors described above with the ``likes''-based measure. The quality of the assessment of an advertisement's success is undoubtedly the best judged by a detailed user study where many of these confounding factors can be controlled for; however, this is expensive and often not feasible to perform each time a new advertisement needs to be appraised.
 
We do note the following interesting trend as it relates to selecting the best model. Regardless of which measure is adopted, the performance trends are near identical.
The multimodal LSTM significantly outperforms the other models, followed by the multimodal DBM, and then the SVM and logistic regression classifiers. This rank order of algorithmic performance in terms of their accuracy and F1-score is the same no matter which measure is used as ground truth. The fact that text features alone are able to achieve a higher performance than the visual and auditory features individually is also corroborated by all the measures. Hence, for the purpose of evaluation and model selection we hypothesize that {\it one can employ metrics derived from easily available online information such as likes, views and comments received by advertisements}, rather than opting for the often much more expensive method of performing a user study.
 
Having said this, we note that using the most expensive method of a user study as ground truth gives a statistically significant improvement in performance over the other measures. For instance, the difference in accuracy between the LSTM evaluated on the user study and on the ratio of likes per visits has a p-value of 0.04123 whereas the difference in F1-score between the two has a p-value of 0.0133, using a McNemar's paired test of significance. Both these values are considered significant.

\smallskip
\subsubsection{Multimedia Feature Analysis}
We additionally seek to study which multimedia attributes seem to contribute the most to the success of commercial advertisements. For this purpose we experimented on three kinds of classifiers: Random Forests~\cite{breiman2001random}, Extra Trees~\cite{geurts2006extremely} and Decision Trees~\cite{breiman1984classification}, to select important video and audio features, that can contribute to a high classification accuracy and are the most responsible in distinguishing between the two classes of advertisements. We obtained a classification accuracy in the range of $0.45-0.55$ with these classifiers. We show the top essential video and audio features that we obtained via the Random Forest classifier in Table~\ref{tab:result:rft:10importance}. The Extra Trees and Decision Tree classifier identified similar if not identical features as important.

\begin{table}[!t]
\caption{Top 10 important video and audio features obtained using a Random Forests classifier}
%\vspace{-0.6em}
\label{tab:result:rft:10importance}
\small
\begin{tabularx}{\linewidth}{L{1}L{1}}
\hline
{\bf  Video Feature} & {\bf Audio Feature}  \\
\hline Variation of intensity span across temporal segments
& Timbre width for the third partition of audio  \\
Variation of average saturation across spatial zones
& Dynamic range for the fourth partition of audio \\
Average saturation span for the fifth spatial zone
&  Onset spectrum strength for the second partition of audio \\
Average chroma for the first video segment
&  Onset spectrum strength for the fourth partition of audio \\
Average intensity span for the third spatial zone  
& Onset spectrum variation for the fourth partition of audio \\
Average intensity for the first video segment
& Onset density for the fourth partition of audio  \\       
Variation of intensity span across spatial zones
 & Timbre width for the fifth partition of audio  \\
Variation of chroma span across temporal segments
 & Mean of timbre width variation  \\
 Average intensity for the second video segment
 & Dynamic range for the second partition of audio \\
Average chroma span for the fifth spatial zone
 & Onset spectrum strength for the third partition of audio \\
\hline
\end{tabularx}
\end{table}
 
In case of the visual attributes, the average intensity and average chroma for the first and second video segments are found to be important. This is understandable because the beginning of the advertisements plays a crucial role in customers deciding whether they want to continue watching it or not. Also features that have been recognized as essential are the average saturation span and average chroma span for the fifth spatial zone, which is the central zone of the screen. This can be because customers pay more attention to the central area of the screen. In case of audio features, we obtain the onset spectrum strength and dynamic range for the second audio partition, which once again showcases the importance of attracting customers at the beginning of the advertisement. Once customers have started watching, the characteristics of the products introduced in the advertisements are reflected by the audio, which are important for an effective advertisement. Thus audio features such as onset spectrum strength, onset spectrum variation, onset density dynamic range are considered important for the consequent and final audio partitions.
 
In order to validate the importance of the above features with respect to our model, we performed experiments using our proposed model, after excluding these particular audio-visual features from the input and using the user study as a ground truth metric. The textual input remained the same as earlier. We found a significant reduction in classification accuracy for the LSTM based model, down to about 67\%, while the accuracy of the DBM based model went down to about 61\%. We then used just the top important audio-visual features and the entire textual feature set as part of the input data, to classify the advertisements with the DBM and LSTM based models. The accuracy of both models was found to reduce to 70\% and 63\% respectively, with the reduction also possibly due to loss of information via feature elimination. However, using only the important features we are still able to manage a reasonable classification accuracy. Thus, we can see that the above identified video and audio features are indeed essential in distinguishing between effective and ineffective advertisements, and can characterize the advertisements well.

\smallskip
\subsubsection{Analysis of Emotional Content}
 
%The results for the LSTM-based emotion detection algorithm seemed to be a little ambiguous, and hence the result for dictionary based emotion detection results is reported in this section. 
We first divided all advertisements in our dataset into categories based on the topic they were about, and then identified the dominant emotion present in their linguistic content. As one would expect, the language of most advertisements seems to echo positive emotions of joy or surprise. But, there were also some that exhibited negative emotions such as sadness or fear, especially those in the categories of medicine, news, and movie trailers. We further sought to understand the emotions invoked in users while viewing the advertisements, and their correlation with the dominant emotion identified in the advertisement language. Table~\ref{tab:emoresult} shows a comparison between the dominant emotions detected in the advertisement text of several categories, and those invoked in the viewers of these advertisements, using the LSTM-based approach for emotion detection (Section~\ref{sec:emotiondetection}). 

According to our emotion detection algorithm most advertisements irrespective of what they advertise, reflect positive emotions such as joy, anticipation, trust and surprise in their content. In general, they inspire feelings of attentiveness, joy, enthusiasm and excitement among their viewers. A good fraction of viewer responses are also neutral towards some product categories such as footwear, clothing and cars. For the most part, the only advertisement categories that invoked negative feelings such as distress, depression, worry or sadness were those that themselves contained such emotions in their content, such as news and medical ailment related commercials, with the exception of movie trailers. There was little correlation between emotions perceived in movie trailer advertisements and user emotions. These primarily invoked a positive affect of joy, enthusiasm or anticipation in viewers, with the exception of less popular movies. We also obtained some non-intuitive finds. For instance, contrary to one’s expectations, we found that an advertisement on the topic `funeral' was detected to invoke a dominant emotion of joy in its viewers, primarily because the language used in the advertisement itself was quite positive with several euphemisms. Interestingly, advertisements where popular restaurants advertised discounts consisted of terms representing a good distribution happiness, anticipation and trust seemed to invoke negative emotions such as anger and irritability in a good fraction of users, though these emotions did not dominate. In general, advertisements shorter in length with lesser content, and without a brand name tended to invoke weaker positive or neutral feelings in users despite a heavy presence of positive affect detected in them. This was irrespective of the kind of product they advertised. 
 
One interesting and important outcome was the emotion `anger' aroused towards different advertisements which was something unexpected. After closely analyzing the video advertisements in question, we discovered that anger was primarily detected in conjunction with those advertisements where products of two different brands were compared and one brand seemed to be belittled. There were also some advertisements with a discrepancy in their video content and text content, primarily seen in advertisements which are `fake' or lack authentic sources and have been generated or uploaded by unregistered users on YouTube. %Our algorithm appears to detect these as well.

Seeing that emotion can play a crucial role in advertisement effectiveness, we sought to include it as an additional feature in our LSTM/DBM model. However since we did not obtain significant performance gains, we do not report these results. 

%\begin{comment}
\begin{table}[!t]
\caption{Comparison between emotions detected in advertisement text and those invoked in viewers, aggregated over selected advertisement topic categories. Neutral represents lack of emotion.}
%\vspace{-0.6em}
\label{tab:emoresult}
\small
\begin{tabularx}{\linewidth}{L{0.4}L{0.8}L{0.8}}
\hline
{\bf Topic} & {\bf Dominant Emotions (ad text)} & {\bf Dominant Emotions (user study)} \\
\hline
Cars & Anticipation, fear & Enthusiasm, excitement, neutral  \\
%\hline
Medical & Sadness, fear, trust & Distress, depression, comfort \\
%\hline
Phones & Anticipation, joy  & Excitement, happiness, influential \\
%\hline
Food/drink & Surprise, joy & Attentiveness, neutral \\
Shoes & Trust, joy & Attentiveness, joy, neutral \\
Clothing  & Joy, surprise & Excitement, neutral \\
Electronics  & Joy, trust, anticipation & Joy, influential \\
Movie trailers  & Surprise, sadness, disgust & Enthusiasm, entertainment, influential \\
News & Sadness, trust, anger & Fear, worry, sadness \\
Finance & Fear, trust, anger & Attentiveness, inspirational \\
\hline
\end{tabularx}
\end{table}
%\end{comment}

\smallskip
\section{Discussion}
Research on how advertising works at the individual level~\cite{vakratsas1999} shows that viewers pay very little attention to details while watching advertisements. This leads to a type of cognitive process called low involvement information processing~\cite{hawkins1992}. Its implication for advertising is that advertisements must first attract the target viewer’s attention before delivering its main message. This implication is in line with our finding that the video segments in the first few seconds of the advertisement significantly marks out effective advertisements from ineffective ones (from Table~\ref{tab:result:rft:10importance}).

Our discovery that audio features of the second audio partition predicts advertisement effectiveness (from Table~\ref{tab:result:rft:10importance}) resonates with results in the marketing literature which reports that message: relevant music can grab viewers' attention~\cite{bruner1990}. However what is new in our results is that the order in which the visual elements and auditory elements occur in an advertisement is what attracts and holds viewers' attention. Our finding that the video features of the central part of an advertisement is important for its effectiveness suggests that this is where the core message of the advertisement is embedded. We also find that brand mentions in an advertisement makes it more effective. This result finds support in the branding literature which confirms that effective advertisements are those that clearly communicate brand benefits~\cite{campbell2003}. However, our finding that the temporal location of brand-mention is irrelevant is not supported by marketing literature. Scholars report that it is better to convey brand names and logos after the introductory attention-grabbing phase. We offer two reasons for our results. First, we do not control for the product category effect. For example, the temporal location of brand mention may vary whether the advertisement is conveying information about a drug or a Disney vacation. Second, we do not control for the brand’s stage in its product life cycle. For example, where the brand name and logo appears may matter differently if we are dealing with a new brand or a mature brand. Our small sample size did not permit this analysis.

Our most intriguing finding is that on average, text features explain advertisement-effectiveness more than either audio (music, loudness, timbre) or video features (see Table~\ref{tab:result}). Recall, that our text features come from the transcription of the voice over in the advertisement. This means that viewers prefer advertisements in which brand benefits are conveyed verbally and with which they can connect emotionally (also see Table~\ref{tab:emoresult}). We know from the information processing literature that our brains process visual, auditory and verbal information through different neural pathways~\cite{keller1998,maclnnis1989}. Though visual and auditory information grab our attention quickly, verbal information is cognitively demanding because when we hear words we try to extract meaning from them~\cite{keller1998}. Verbally presented brand information  often forms strong associations in our neural networks~\cite{keller1998}.

Taken together our findings show that effective advertisements on social media websites like YouTube, involve the blending of visual, auditory and linguistic features. Creating effective advertisements requires the careful sequencing of these features so that the advertisement first draws the viewer's attention and then drives home the brand message. How the advertisement begins is critical to its effectiveness, since it is the beginning that sets off the narrative arc of the story that is contained in the remainder of the advertisement. As with any good story without an interesting beginning the reader rarely moves on to finish the story (Table~\ref{tab:result:rft:10importance}).
\section{Conclusion}
\label{sec:conclusion}
 
In this work, we analyze various temporal features and multimedia content of online advertisement videos which include auditory, visual and textual components and study their role in advertisement success. We trained three individual neural network models employing features from the modalities of video, audio and text, and fused together the resultant representations from each of these to learn a joint embedding. We applied this joint embedding to a binary softmax classifier to predict advertisement effectiveness or success. The performance of our approach was validated on subjective grounds based on a user study, the sentiment strength of user comments and the ratio of likes to visits from YouTube on the respective advertisements. The unusual effectiveness and lift obtained from the novel LSTM variants we propose (over strong baselines) on this challenging problem seem in line with recent efforts in the image processing, computer vision and language modeling domains~\cite{venugopalan2016,gulcehre2015}.
In future, we would like to automate our feature engineering process utilizing CNNs trained on datasets such as ImageNet, and also try to understand and interpret the decision process implicit in such models, to provide direct recommendations to advertisers and marketers. We are also interested in developing models capable of providing a more fine-grained outcome for an advertisement rather than just a binary value of `effective' or `ineffective' to provide additional insights on its effectiveness.

\section*{Acknowledgments}
This work is supported by NSF awards NSF-EAR-1520870 and NSF-SMA-1747631.

%\section*{References}

%\newpage
\bibliographystyle{IEEEtran}
\bibliography{IEEEabrv,references}

\end{document}